\documentclass{article}
\usepackage{spconf,amsmath,graphicx}
\usepackage{url}
\usepackage{multirow}
\usepackage{caption}
\usepackage{rotating}
\usepackage{enumitem}  
\usepackage{lipsum}
\usepackage[hidelinks]{hyperref}
\usepackage{multirow}

\title{The Broad Impact of Feature Imitation: Neural Enhancements Across Financial, Speech, and Physiological Domains}

\name{Reza Khanmohammadi$^{\dagger}$ \; Tuka Alhanai$^{\mathsection}$ \; \textit{Mohammad M. Ghassemi}$^{\dagger}$}
\address{$^{\dagger}$Computer Science and Engineering Department, Michigan State University \\
$^{\mathsection}$Department Computer Engineering, New York University Abu Dhabi}

\begin{document}

\maketitle

\begin{abstract}
Initialization of neural network weights plays a pivotal role in determining their performance. Feature Imitating Networks (FINs) offer a novel strategy by initializing weights to approximate specific closed-form statistical features, setting a promising foundation for deep learning architectures. While the applicability of FINs has been chiefly tested in biomedical domains, this study extends its exploration into other time series datasets. Three different experiments are conducted in this study to test the applicability of imitating Tsallis entropy for performance enhancement: Bitcoin price prediction, speech emotion recognition, and chronic neck pain detection. For the Bitcoin price prediction, models embedded with FINs reduced the root mean square error by around 1000 compared to the baseline. In the speech emotion recognition task, the FIN-augmented model increased classification accuracy by over 3 percent. Lastly, in the CNP detection experiment, an improvement of about 7 percent was observed compared to established classifiers. These findings validate the broad utility and potency of FINs in diverse applications.
\end{abstract}

\begin{keywords}
Feature Imitating Network, Bitcoin Price Prediction, Speech Emotion Recognition, Chronic Neck Pain  
\end{keywords}

\section{Introduction}
\label{sec:intro}

Deep learning has established itself as a foundational technique across various applications, primarily due to its capability to learn complex patterns and relationships. One of the crucial aspects influencing the efficacy of deep learning models is the initialization of their weights. Proper weight initialization can lead to faster model convergence and enhanced performance \cite{kumar2017weight}. While the reliance on large datasets and extensive computational resources is vital for determining feature quality and model versatility, correct initialization can offset some of the dependencies on these resources. This offset is especially crucial in domains with limited data and computational capabilities, underlining the importance of leveraging deep learning's potential without a heavy reliance on large datasets and extensive resources. To cater to such scenarios, FINs \cite{9746397} offer an intuitive approach where neural networks are initialized to imitate specific statistical properties. By doing so, FINs provide a more informed starting point, making neural networks less opaque and offering a hint of interpretability in what is often dubbed a "black box." The beauty of FINs lies in their simplicity, allowing researchers to directly incorporate domain-specific knowledge into the model's architecture, fostering both efficacy and understandability.

\subsection{Contributions}
While FINs have made significant strides in biomedical signal processing \cite{9746397,min2023feature,khanmohammadi2022mambanet}, their applicability in broader domains remains a topic of interest. In this work, we delve into the potential of FINs across three distinct areas: financial, speech, and Electromyography (EMG) time series analysis. Our research aims to demonstrate how integrating a lightweight FIN can enhance the performance of different neural network architectures, regardless of the task or network topology. By investigating their effects across different contexts, we offer insights into the adaptability, benefits, and potential boundaries of using FINs.

\section{Related Work}
\label{sec:relatedwork}
\textbf{The Evolution of Transfer Learning Across Domains:} Transfer learning has emerged as a potent technique in machine learning, reshaping the paradigm by repurposing pre-trained models to tackle different tasks from their original intent \cite{zhuang2020comprehensive}. Such a strategy has yielded transformative advancements, especially in computer vision \cite{LI2020103853}, speech analysis \cite{amiriparian2021deepspectrumlite}, and natural language processing (NLP) \cite{ruder-etal-2019-transfer}. Foundational models like ResNet \cite{he2015deep}, wav2vec \cite{schneider2019wav2vec}, and BERT \cite{devlin2019bert} stand as prime examples of this shift, requiring significantly reduced training data when finetuned for new tasks. Transitioning this approach to the biomedical arena presents unique challenges. There is an inherent lack of large and diverse biomedical datasets [10]\cite{Ghassemi2018YouSY}, which has led to cross-domain adaptations, such as repurposing computer vision models for audio classification \cite{Lech2020RealTimeSE}. These adaptations, while novel, often do not achieve the same efficacy as within-domain counterparts, highlighting the pressing need for tailored approaches for biomedical data.

\noindent\textbf{Statistical Feature Imitation Bridges the Transfer Learning Divide in Diverse Specialized Tasks:} FINs have established a unique role in addressing this particular challenge \cite{9746397}. FINs offer a distinctive approach to neural learning by initializing weights to simulate distinct statistical features, effectively bridging domain knowledge with machine learning. This method has catalyzed notable progress in many fields by showcasing its effectiveness across various tasks. 
In the seminal work introducing FINs \cite{9746397}, the authors showcased the efficacy of this novel approach across three experiments. In Electrocardiogram (ECG) artifact detection, a FIN imitating the Kurtosis feature outperforms standard models in both performance and stability. Similarly, for Electroencephalogram (EEG) artifact detection within the same research, FINs imitating Kurtosis and Shannon's Entropy enhanced results. Moreover, when applied to EEG data for fatigue and drowsiness detection, a FIN based on Shannon's entropy consistently outperformed baselines, while certain models like VGG proved ineffective. Additionally, FINs have shown promise in specialized applications. In biomedical image processing, Ming et al (2023) provided state-of-the-art results across tasks including COVID-19 detection from CT scans and brain tumor identification and segmentation from MRI scans \cite{min2023feature}. In sports analytics, the hybrid architecture of MambaNet \cite{khanmohammadi2022mambanet} employed FINs to effectively predict NBA playoff outcomes, showcasing the broad versatility of the FIN approach. Although FINs have shown promise in biomedical applications and sports analytics, their potential in financial and speech time series data is yet to be explored.

\vspace{-5pt}
\section{Imitating Tsallis Entropy}
\label{sec:methods}
A FIN is a neural network that is trained to approximate a closed-form statistical feature of choice. In our study, we train a FIN to imitate Tsallis Entropy. Tsallis entropy, a non-extensive generalization of the traditional Shannon entropy, measures the uncertainty or randomness of a system. Uniquely, it takes into account the correlations and higher-order interactions that are often overlooked by the conventional Shannon entropy. This quality makes Tsallis entropy particularly apt for systems exhibiting non-standard statistical behaviors and long-range dependencies.

\noindent\textbf{The Influence of $q$ on Tsallis Entropy} \quad The distinguishing characteristic of Tsallis entropy is its reliance on the parameter $q$. The Shannon entropy becomes a special case of Tsallis entropy when $q=1$. When $q>1$, the entropy gives more weight to lower probabilities, making it more sensitive to rare events. Conversely, for $q<1$, the entropy calculation is dominated by higher probabilities. This variability in weighting is encapsulated by the equation for a discrete probability distribution $p(i)$ as influenced by the temperature scaling parameter $\tau$:
\begin{equation}
H_q(\tau) = \frac{1}{q - 1} \left(1 - \sum_{i} \text{softmax}\left(\frac{u(i)}{\tau}\right)^q\right)
\label{eq:tsallis}
\end{equation}
Where \( u(i) \) represents the unscaled probabilities from the normalized input. In our implementation, $q$ is set to a default value of 1.5 and further treated as a trainable parameter within our FIN, allowing the model to adaptively finetune its value to optimally capture the inherent complexities and nuances of the dataset.

\noindent\textbf{Temperature Scaling with Parameter $\tau$} \quad Another pivotal parameter in our approximation process is $\tau$. This temperature parameter modulates the entropy's sensitivity by scaling the inputs to the softmax function. Specifically, as $\tau$ approaches 0, the softmax output mirrors a one-hot encoded distribution, while increasing $\tau$ causes the resultant distribution to edge towards uniformity. The introduction of $\tau$ in the Tsallis entropy equation underlines its importance in shaping the final probabilities. In the context of our work, $\tau$ is initialized with a default value of 1, but like $q$, it's also trainable within our FIN, allowing the network to adjust it adaptively during the learning phase.

\noindent\textbf{Training} \quad To approximate the Tsallis entropy using neural networks, we generated synthetic signals with uniform random values between 0 and 1. The output regression values for the FIN were the Tsallis Entropy values, which were computed directly on the synthetic signals using the defined closed-form expression in equation \ref{eq:tsallis}. This calculation is fundamentally based on a power-law probability distribution.  We utilized a simple gradient descent optimizer along with mean absolute error (L1) loss to train this network. Additionally, early stopping was integrated, and the training was optimized with learning rate modifications facilitated by the \texttt{ReduceLROnPlateau} scheduler.

\noindent\textbf{Baseline Model} \quad In each of our three experiments, we employed a neural network as a comparative baseline. This network had a representational capability (i.e. number of parameters) that was either equal to or exceeded the FIN-embedded networks introduced in that particular experiment. We investigated multiple network topologies, experimenting with as many as ten variations for each baseline. The model that showcased the best performance on the validation set was subsequently chosen for comparison against the Tsallis Entropy FIN-powered networks.

\vspace{-10pt}
\section{Experiments \& Results}
\label{sec:experiments}

\begin{table}
\small
\centering
\caption{Comparative evaluation of models using RMSE and MAPE metrics over the two periods.}
\label{tab:exp1}
\begin{tabular}{|c|c|c|c|}
\hline
\multicolumn{1}{|c|}{\textbf{Period}} & \multicolumn{1}{c|}{\textbf{Model}} & \textbf{RMSE} & \textbf{MAPE} \\
\hline
\multirow{2}{*}{1} & RF regression & 321.61 & 3.39\% \\
& Deep LSTM & 330.26 & 3.57\% \\
& Deep LSTM + Attention & 283.83 & 2.97\% \\
& NN-Baseline & 287.47 & 2.97\% \\
& FIN-ENN & \textbf{277.45} & \textbf{2.87\%} \\
\hline
\multirow{2}{*}{2} & RF regression & 2096.24 & 3.29\% \\
& Deep LSTM & 3045.87 & 4.68\% \\
& Deep LSTM + Attention & 2014.43 & 2.96\% \\
& NN-Baseline & 2127.70 & 3.18\% \\
& FIN-ENN & \textbf{2001.45} & \textbf{2.96\%} \\
\hline
\end{tabular}
\vspace{-10pt}
\end{table}

\subsection{Experiment \Roman{subsection}}  

\noindent\textbf{Objective} \quad This experiment focuses on predicting the closing price of Bitcoin on a given day, for the subsequent day. We hypothesize that we can achieve enhanced predictive accuracy over traditional baselines by initializing certain neural network weights to imitate Tsallis entropy, followed by fine-tuning during training.

\noindent\textbf{Data and Preprocessing} \quad Our study leveraged a publicly accessible dataset\footnote{\urlstyle{same}\url{https://github.com/shiitake-github/jrfm-2156907-data}} that spanned over seven years, from March 2015 to April 20221. Owing to notable Bitcoin price fluctuations in 2017 and 2021, the dataset was bifurcated into two periods: Period 1, from March 2015 to September 2018, and Period 2, from October 2018 to April 2022. Each period was split into approximately an 85 to 15\% ratio for training and testing. The dataset encompassed a total of 47 features clustered into various categories such as Bitcoin price variables, specific technical features of Bitcoin, other cryptocurrencies, commodities, market indexes, and more. In the original study conducted by Chen (2023) \cite{jrfm16010051}, ten features were utilized for Period 1: Bitcoin's opening price, highest price of the day, lowest price of the day, closing price, price of one Ethereum in USD, WTI crude oil price, the Standard and Poor’s 500, National Association of Securities Dealers Automated Quotations (NASDAQ), Dow Jones Industrial Average (DJI), and the mean difficulty on a given day of finding a hash that meets the protocol-designated requirement. For Period 2, a subset of six features was used: the first four Bitcoin-specific prices, the price of one Ethereum in USD, and the Nikkei 225, determined through feature selection. In contrast, our study consistently employed this subset of six features across both periods, as it led to improved results.

\noindent\textbf{Methods} \quad While the baseline architectures include a Random Forest (RF) regression and a deep LSTM network \cite{jrfm16010051}, our research takes this foundation a step further. We introduce a new model, namely Deep LSTM + Attention, which is inspired by the LSTM's structural elements but incorporates significant advancements. Contrary to the original RF regression and LSTM models, our design integrates the last seven timesteps of each feature, enriching its grasp on historical data and potentially enhancing its predictive prowess. Moreover, we incorporated two distinct attention mechanisms: one at the input level and another within the network layers, aiming for refined data representations. Complementing these improvements, we embedded and fine-tuned the Tsallis entropy FIN within this network (FIN-ENN), serving as a transformative layer to delve deeper into the financial intricacies.

\noindent\textbf{Results} \quad The results of our analysis can be found in Table \ref{tab:exp1}. We used Root Mean Square Error (RMSE) that gauges prediction deviations from actual values, and Mean Absolute Percentage Error (MAPE) which quantifies relative error in percentage terms as two metrics for evaluating model performance. In our investigation, we discovered that our introduced Attention-based LSTM network outperformed both the RF regression and LSTM models from the original baseline study. Our model's improvement over the baseline can be attributed to refined neural modeling. Notably, this improvement can be attributed to the meticulous integration of the attention mechanism and extended window size, capturing the last seven timesteps as opposed to the 1 and 2 timestep windows in the original work. Our results indicate a clear superiority of the longer window in effectively predicting next-day closing prices. Building on this, the FIN-Embedded Neural Network (FIN-ENN), which embeds Tsallis entropy at the input level, showcased even greater performance. Specifically, it further decreased prediction errors by 44.16 RMSE and 0.52 MAPE in Period 1, and 94.79 RMSE and 0.33 MAPE in Period 2 when compared to the baseline. The Tsallis entropy is evidently a significant factor in price prediction, as illustrated by our final model. By leveraging this entropy, we've effectively harnessed the temporal intricacies of the financial dataset, thus ensuring more precise forecasts.

\vspace{-5pt}
\subsection{Experiment \Roman{subsection}}

\noindent\textbf{Objective} \quad This experiment aims to enhance speech emotion recognition by leveraging FINs. Unlike the previous experiment, where the input data was fed directly into the FIN, here, we utilize a latent representation of the data—a condensed, yet informative, representation derived from previous layers of a deep neural network. Our hypothesis posits that by feeding this latent representation through the FIN, specifically designed to imitate the Tsallis entropy, and further fine-tuning it during training, we can achieve superior recognition performance. Our target is to surpass the state-of-the-art (SOTA) model, the Acoustic CNN (1D) from the reference study.

\begin{table}
\small
\centering
\caption{Comparison of emotion recognition accuracy across different models and input features.}
\label{tab:exp2}
\begin{tabular}{|c|c|c|}
\hline
\textbf{Method} & \textbf{Input Feature} & \textbf{Accuracy} \\
\hline
Acoustic CNN (1D) & emo\_large & 66.12 \\
\hline
NN-Baseline & w2v2-persian-v3 & 69.40 \\
FIN-ENN & w2v2-persian-v3 & \textbf{72.23} \\
\hline
NN-Baseline & w2v2-persian-ser & 94.87 \\
FIN-ENN & w2v2-persian-ser & \textbf{95.51} \\
\hline
\end{tabular}
\vspace{-10pt}
\end{table}

\noindent\textbf{Data and Preprocessing} \quad We used the publicly available modified version\footnote{\urlstyle{same}\url{https://github.com/aliyzd95/modified_shemo}} of the Sharif Emotional Speech Database (ShEMO) \cite{shemo}, which contains 3 hours and 25 minutes of semi-natural speech samples in .wav format. These 3000 samples, recorded by 87 native Farsi speakers, are multi-labeled. The reference study \cite{shemoModified} concentrated on emotions like anger, happiness, sadness, surprise, and neutral. Each speech segment, with an average length of 4.11 seconds, was embedded using wav2vec2 \cite{w2v2} to enhance its representation in our neural network model.

\noindent\textbf{Methods} \quad Our method is a deep neural network with a series of fully connected (dense) layers with decreasing units: 512, 256, 128, 64, and 32. Each layer is followed by a ReLU activation function and a dropout layer (rate=0.5) to prevent overfitting. Crucially, after obtaining the 32-unit latent representation from the penultimate layer, the FIN is integrated to compute the Tsallis entropy of this representation. The computed entropy is then concatenated with the 32-unit latent representation and fed into the final fully connected layer to produce the output corresponding to the emotion classes. 

\noindent\textbf{Results} \quad Our experiment compared three models: our proposed FIN-ENN, the NN-Baseline, and the Acoustic CNN (1D) from the reference study \cite{shemoModified}. The baseline model utilized the emo\_large feature set, extracting 6552 high-level acoustic features from each audio file using the openSMILE toolkit \cite{opensmile}. These features arise from high-level statistical functions applied to low-level descriptors. Conversely, our FIN-ENN model adopted two fine-tuned versions of the wav2vec2 model: w2v2-persian-v3 \footnote{\urlstyle{same}\url{https://huggingface.co/m3hrdadfi/wav2vec2-large-xlsr-persian-v3}} and w2v2-persian-ser\footnote{https://huggingface.co/m3hrdadfi/wav2vec2-xlsr-persian-speech-emotion-recognition}. As shown in Table \ref{tab:exp2}, the FIN-ENN model's integration of Tsallis FIN contributed to an absolute accuracy improvement of 2.83\% for w2v2-persian-v3 and 0.64\% for w2v2-persian-ser compared to their FIN-less counterparts.

\vspace{-5pt}
\subsection{Experiment \Roman{subsection}}

\noindent\textbf{Objective} \quad This experiment delves into the detection of Chronic Neck Pain (CNP) through EMG data. We hypothesize that embedding a neural network with the FIN, specifically designed to imitate the Tsallis entropy, will improve CNP detection performance compared to traditional models.

\begin{table}
\small
\centering
\caption{Comparison of classification performance in CNP detection.}
\label{tab:exp3}
\begin{tabular}{|c|c|c|c|}
\hline
\textbf{Method} & \textbf{Accuracy} & \textbf{Specificity} & \textbf{Sensitivity} \\
\hline
K-NN (raw) & 35.00 & 35.00 & 35.00 \\
SVM (raw) & 32.50 & 31.57 & 33.33 \\
LDA (raw) & 42.50 & 42.85 & 42.10 \\
\hline
K-NN (NCA) & 55.00 & 54.54 & 55.55 \\
SVM (NCA) & 55.00 & 60.00 & 54.17 \\
LDA (NCA) & 55.00 & 56.25 & 55.00 \\
\hline
NN-Baseline (raw) & 57.50 & 55.00 & 60.00 \\
FIN-ENN (raw) & \textbf{62.50} & \textbf{65.00} & \textbf{60.00} \\
\hline
\end{tabular}
\end{table}

\noindent\textbf{Data and Preprocessing} \quad Our dataset, sourced from Jim{\'e}nez-Grand et al \cite{jimenez2021muscle} and publicly available on Kaggle\footnote{\urlstyle{same}\url{https://www.kaggle.com/datasets/david893/neck-emg-data}}, consists of twenty asymptomatic individuals and twenty with Chronic Neck Pain (CNP). Asymptomatic individuals had no significant neck issues in the last two years, while CNP individuals reported notable pain recently. Data was collected as participants walked barefoot along a six-meter rectilinear path, repeated three times at two-minute intervals. Building upon the approach adopted in the original study by Jim{'e}nez-Grand et al. \cite{jimenez2021muscle}, we extracted the same four time domain and six frequency domain features from the EMG data. However, instead of analyzing every 500 ms of the signal (as determined by a 1000Hz sampling rate), we segmented the entire signal into five distinct parts, a method inspired by \cite{9848340}. Similarly to prior studies, our focus centered on four upper back muscle groups: Trapezius, Sternocleidomastoid, C4 Paraspinal, and Latissimus Dorsi, with each muscle group including both left and right muscles, and features were computed for each side.

\noindent\textbf{Methods} \quad Jim{'e}nez-Grand et al. \cite{jimenez2021muscle} employed K-NN, SVM, and LDA for classification, processing both raw and Neighbourhood Component Analysis (NCA)-selected features \cite{nca}. In contrast, we used the raw extracted features to train a feed-forward neural network comprising two hidden layers with 256 and 32 units. Drawing inspiration from our previous experiment, the 32-dimensional latent representation from the second hidden layer was channeled into the Tsallis FIN. This processed output was then concatenated with the original 32 features, yielding a 33-dimensional vector that was finally directed to a sigmoid activation to perform the binary classification.

\noindent\textbf{Results} \quad As outlined in Table \ref{tab:exp3}, we compared the performance of our FIN-ENN model against those developed in the original study using accuracy, specificity, and sensitivity. The original study's models, namely K-NN, SVM, and LDA, achieved a maximum accuracy of 55.00\% with NCA-selected features. Our NN-Baseline registered an accuracy of 57.50\%. However, by leveraging the Tsallis FIN in our architecture, we achieved a superior accuracy of 62.50\%. This improvement is also evident in improvements made in both specificity (65.00\%) and sensitivity (60.00\%). Our results reinforce our initial hypothesis, underscoring the benefits of incorporating the FIN for CNP detection from physiological EMG data.

\vspace{-10pt}
\section{Conclusion}
\label{sec:conclusion}
In our experiments, integrating a Feature Imitating Network (FIN) designed to imitate Tsallis entropy consistently enhanced predictive model performances across diverse domains. In predicting Bitcoin's subsequent day's closing price, the enhanced neural network outshone traditional models like Random Forest regression and LSTM. Similarly, in speech emotion recognition, the FIN-augmented model excelled at processing latent representations. In detecting Chronic Neck Pain (CNP) through EMG data, it surpassed established classifiers like K-NN, SVM, and LDA. The consistent edge the FIN provides across these areas underscores its broad utility and efficacy. Future studies can more profoundly investigate influential financial, speech, and physiological features to imitate, aiming to amplify the performance of neural predictive models further.

\bibliographystyle{IEEEbib}
\bibliography{refs}
\end{document}